\documentclass[12pt]{article}
\usepackage{amsbsy,amssymb,amsmath,amsthm,amscd,amsfonts,latexsym,amstext,delarray,
amsmath,graphicx} 
\usepackage[margin=1in]{geometry}
\usepackage{color}

\newcommand{\F}{{\mathbb F}}
\newcommand{\Q}{{\mathbb Q}}
\newcommand{\cV}{{\mathcal V}}
\newcommand{\cU}{{\mathcal U}}

\newtheorem{ex}{Example}

\numberwithin{equation}{section}

\title{Principles and Parameters: a coding theory perspective}

\author{Matilde Marcolli \\
  Department of Mathematics, Caltech \\
  1200 E. California Boulevard, Mail Code 253-37 \\
  Pasadena, CA 91125, USA \\
  {\tt matilde@caltech.edu}\\ }
  
  \date{}

\begin{document}
\maketitle

\begin{abstract}
We propose an approach to Longobardi's parametric comparison method (PCM) via the theory
of error-correcting codes. One associates to a collection of languages to be analyzed with the
PCM a binary (or ternary) code with one code words for each language in the family and
each word consisting of the binary values of the syntactic parameters of the language, 
with the ternary case allowing for an additional parameter state that takes into account 
phenomena of entailment of parameters. The code parameters of the resulting code can
be compared with some classical bounds in coding theory: the asymptotic bound, the 
Gilbert--Varshamov bound, etc. The position of the code parameters with respect to
some of these bounds provides quantitative information on the variability of syntactic
parameters within and across historical-linguistic families. While computations carried out
for languages belonging to the same family yield codes below the GV curve, comparisons
across different historical families can give examples of isolated codes lying 
above the asymptotic bound.
\end{abstract}

\section{Introduction}

The generative approach to linguistics relies on the notion of a Universal Grammar (UG) 
and a related universal list of syntactic parameters. In the Principles
and Parameters model, developed since \cite{Chom}, 
these are thought of as binary valued parameters or ``switches"
that set the grammatical structure of a given language. Their universality makes it
possible to obtain comparisons, at the syntactic level, between arbitrary pairs of natural languages. 

A parametric comparison method (PCM) was introduced in \cite{Longo2} as a
quantitative method in historical linguistics, for comparison of languages within
and across historical families at the syntactic instead of the lexical level. Evidence 
was given in \cite{LoGua} and \cite{LGSBC} that the PCM gives reliable information
on the phylogenetic tree of the family of Indo--European languages.

The PCM relies essentially on constructing a metric on a family of languages
based on the relative Hamming distance between the sets of parameters as
a measure of relatedness. The phylogenetic tree is then constructed on the
basis of this datum of relative distances, see \cite{LoGua}.

Our purpose in this paper is to connect the PCM approach to the mathematical
theory of error-correcting codes. We associate a code to any group of languages one
wishes to analyze via the PCM, which has one code word for each language. 
If one uses a number $n$ of syntactic parameters, then the code $C$ sits in the
space $\F_2^n$, where the elements of $\F_2=\{ 0,1 \}$ correspond to the two $\mp$
possible values of each parameter, and the code word of a language is the string
of values of its $n$ parameters. We also consider a version with codes on an
alphabet $\F_3$ of three letters which allows for the possibility that some of the
parameters may be made irrelevant by entailment from other parameters. In this
case we use the letter $0 \in \F_3$ for the irrelevant parameters and the nonzero
values $\pm 1$ for the parameters that are set in the language. 

In the theory of error-correcting codes, see \cite{TsfaVla}, one assigns to a
code $C\subset \F_q^n$ two code parameters: $R=\log_q(\# C)/n$, the
transmission rate of the code, and $\delta=d/n$ the relative minumum
distance of the code, where $d$ is the miminum Hamming distance between
pairs of distinct code words. It is well known in coding theory that ``good codes"
are those that maximize both parameters, compatibly with several constraints
relating $R$ and $\delta$. In particular, it was proved in \cite{Man} that
there is a curve $R=\alpha_q(\delta)$ in the space of code parameters, the
asymptotic bound, that separates code points that fill a dense region and
that have infinite multiplicity from isolated code points that only have finite
multiplicity. These better but more elusive codes are typically obtained through
algebro-geometric constructions, see \cite{Man}, \cite{TsfaVlaZi}, \cite{VlaDri}.
The asymptotic bound was recently related to Kolmogorov complexity in
\cite{ManMar2}.

Given a collection of languages one wants to compare through their
syntactic parameters, one can ask natural questions about the position
of the resulting code in the space of code parameters and with respect
to the asymptotic bound. The theory of error correcting codes tells us
that codes above the asymptotic bound are very rare, and indeed one
finds that, in all cases we looked at, languages belonging to the
same historical-linguistic family yield codes below the asymptotic
bound (and in fact below the Gilbert--Varshamov curve). This gives 
a precise quantitative bound to the possible spread of syntactic
parameters compared to the size of the family, in terms of the number
of different languages belonging to the same historico-linguistic group.
However, we show that, if one considers sets of languages that do
not belong to the same historical-linguistic family, then one can
obtain codes that lie above the asymptotic bound, a fact that reflects
in this code theoretic terms, the much greater variability of syntactic
parameters. The result is in itself not surprising, but the point we
wish to make is that the theory of error-correcting codes provides 
a natural setting where quantitative statements of this sort can
be made using methods already developed for the different purposes of
coding theory. We conclude by listing some new linguistic questions that arise
by considering the parametric comparison method under this coding
theory perspective.

\section{Language families as codes}

The Principles and Parameters model of Linguistics assigns to every
natural language $L$ a set of binary values parameters that describe
properties of the syntactic structure of the language.

Let $F$ be a {\em language family}, by which we mean a finite collection
$F=\{ L_1, \ldots, L_m \}$ of languages. This may coincide with a
family in the historical sense, such as the Indo-European family,
or a smaller subset of languages related by historic origin and
development (e.g. the Indo-Iranian, or Balto-Svalic languages), or
simply any collection of language one is interested in comparing
at the parametric level, even if they are spread across different
historical families.
 
We denote by $n$ be the number of parameters used in the
parametric comparison method. We do not fix, a priori, a 
value for $n$, and we consider it a variable of the model.
We will discuss below how one views, in our perspective, the issue 
of the independence of parameters.

After fixing an enumeration of the parameters, that is, a bijection
between the set of parameters and the set $\{ 1, \ldots, n \}$, we
associate to a language family $F$ a code $C=C(F)$ in $\F_2^n$,
with one code word for each language $L\in F$, with the code
word $w=w(L)$ given by the list of parameters $w=(x_1, \ldots, x_n)$,
$x_i\in \F_2$ of the language. For simplicity of notation, we just write
$L$ for the word $w(L)$ in the following.

In this model, we only consider binary parameters
with values $\pm 1$ (here identified with letters $0$ or $1$ in $\F_2$)
and we ignore parameters in a neutralized state following 
implications across parameters, as in the datasets of \cite{LGSBC}, \cite{LoGua}.
The entailment of parameters, that is, the phenomenon by which a particular
value of one parameter (but not the complementary value) renders another
parameter irrelevant, was addressed in greater detail in \cite{Longo}.
We discuss a version of our coding theory model that does
not incorporate entailment, but we comment in \S \ref{entail} below
how this can be modified to incorporate this phenomenon.

The idea that natural languages can be described, at the level of their
core grammatical structures, in terms of a string of binary characters (code 
words) was already used extensively in \cite{ClaRo}.

\subsection{Code parameters}

In the theory of error-correcting codes, one assigns two main
parameters to a code $C$, the {\em transmission rate} and the
{\em relative minimum distance}. More precisely, a binary code
$C\subset \F_2^n$ is an $[n,k,d]_2$-code if the number of code
words is $\# C =2^k$, that is,
\begin{equation}\label{kC}
k = \log_2 \# C,
\end{equation}
where $k$ need not be an integer, and the
minimal Hamming distance between code words is
\begin{equation}\label{dC}
d = \min_{L_1\neq L_2 \in C} d_H(L_1,L_2),
\end{equation}
where the Hamming distance is given by 
$$d_H(L_1,L_2) = \sum_{i=1}^n |x_i - y_i|,$$ for $L_1=(x_i)_{i=1}^n$
and $L_2=(y_i)_{i=1}^n$ in $C$. The transmission rate of the code $C$
is given by
\begin{equation}\label{RC}
R = \frac{k}{n} .
\end{equation}
One denotes by $\delta_H(L_1,L_2)$ the
relative Hamming distance
$$ \delta_H(L_1,L_2)= \frac{1}{n} \sum_{i=1}^n |x_i - y_i|, $$
and one defines the relative minimum distance of the code $C$ as
\begin{equation}\label{deltaC}
\delta = \frac{d}{n} = \min_{L_1\neq L_2\in C}  \delta_H(L_1,L_2).
\end{equation}

\smallskip

In coding theory, one would like to construct codes that simultaneously
optimize both parameters $(\delta, R)$: a larger value of $R$ represents
a faster transmission rate (better encoding), and a larger value of $\delta$
represents the fact that code words are sufficiently sparse in the ambient
space $\F_2^n$ (better decoding, with better error-correcting capability).
Constraints on this optimization problem are expressed in the form of
bounds in the space of $(\delta,R)$ parameters, see \cite{Man}, \cite{TsfaVla}.

\smallskip

In our setting, the $R$ parameter measures the ratio between the
logarithmic size of the number of languages being encompassing
the given family and the total number of parameters, or equivalently
how densely the given language family is in the  ambient 
configuration space $\F_2^n$ of parameter possibilities. The parameter
$\delta$ is the minimum, over all pairs of languages in the given
family, of the relative Hamming distance used in the PCM method
of \cite{LGSBC}, \cite{LoGua}.

\subsection{Parameter spoiling}

In the theory of error-correcting codes, one considers {\em spoiling operations}
on the code parameters. Applied to an $[n,k,d]_2$-code $C$, these produce, respectively,
new codes with the following description (see \S 1.1.1 of \cite{ManMar}):
\begin{itemize}
\item A code $C_1=C\star_i f$ in $\F_2^{n+1}$, for a map $f: C \to \F_2$, whose
code words are of the form $(x_1,\ldots, x_{i-1},f(x_1,\ldots,x_n), x_i, \ldots, x_n)$
for $w=(x_1,\ldots,x_n)\in C$. If $f$ is a constant function, $C_1$ is an $[n+1,k,d]_2$-code.
If all pairs $w,w' \in C$ with $d_H(w,w')=d$ have $f(w)\neq f(w')$, then $C_1$ is an
$[n+1,k,d+1]_2$-code.
\item A code $C_2= C\star_i$ in $\F_2^{n-1}$, whose code words are given by the projections
$$(x_1,\ldots, x_{i-1},x_{i+1},\ldots,x_n)$$ of code words $(x_1,\ldots, x_{i-1},x_i,x_{i+1},\ldots,x_n)$
in $C$. This is an $[n-1,k,d-1]_2$-code, except when all pairs $w,w' \in C$ with 
$d_H(w,w')=d$ have the same letter $x_i$, in which case it is an $[n-1,k,d]_2$-code.
\item A code $C_3= C(a,i) \subset C \subset \F_2^n$, given by the level set 
$C(a,i)=\{ w=(x_k)_{k=1}^n \in C \,|\, \, x_i =a \}$. Taking $C(a,i)\star_i$ gives an
$[n-1, k',d']_2$-code with $k-1\leq k'< k$, and $d'\geq d$.
\end{itemize}
The same spoiling operations hold for $q$-ary codes $C\subset \F_q^n$,
for any fixed $q$.

In our setting, where $C$ is the code obtained from a family of languages, according
to the procedure described above, the first spoiling operation can be seen as the
effect of considering one more syntactic parameter, which is dependent on the other
parameters, hence describing a function $F: \F_2^n \to \F_2$, whose restriction to
$C$ gives the function $f: C \to \F_2$. In particular, the case where $f$ is constant on $C$
represents the situation in which the new parameter adds no useful comparison information
for the selected family of languages. The second spoiling operation consists in forgetting one
of the parameters, and the third corresponds to forming subfamilies of the given family
of languages, by grouping together those languages with a set value of one of the
syntactic parameters. Thus, all these spoiling operations have a clear meaning from
the point of view of the linguistic PCM.

\medskip

\subsection{Examples}

We consider the same list of 63 parameters used in \cite{LoGua} (see \S 5.3.1 and Table A).
This choice of parameters follows the {\em modularized global parameterization} 
method of \cite{Longo2}, for the Determiner Phrase module.
They encompass parameters dealing with person, mumber, and gender (1--6 on their list),
parameters of definiteness (7--16 in their list), of countability (17--24), genitive structure (25--31),
adjectival and relative modification (32--14), position and movement of the head noun
(42--50), demonstratives and other determiners (51--50 and 60--63), possessive pronouns (56--59);
see \S\S 5.3.1--5.3.2 of \cite{LoGua} for more details.

Our very simple examples here are just meant to clarify our notation: they
consist of some collections of languages selected from the list of 28,
mostly Indo--European, languages considered in \cite{LoGua}. In each group we 
consider we eliminate the parameters that are entailed from others, and we focus on
a shorter list, among the remaining parameters, that will suffice to illustrate our viewpoint.

\begin{ex}\label{ex1}{\rm Consider a code $C$ formed out of the languages $\ell_1=$
Italian, $\ell_2=$ Spanish, and $\ell_3=$ French, and let us consider only the first six
syntactic parameters of Table A of \cite{LoGua}, so that $C\subset \F_2^n$ with $n=6$.
The code words for the three languages are
{\small
\begin{center}
\begin{tabular}{|c|c|c|c|c|c|c|}
\hline
$\ell_1$ & 1 & 1 & 1 & 0 & 1 & 1 \\
\hline
$\ell_2$ & 1 & 1 & 1 & 1 & 1 & 1 \\
\hline
$\ell_3$ & 1 & 1 & 1 & 0 & 1 & 0 \\
\hline
\end{tabular}
\end{center}
}
}
\end{ex}

This has code parameters $(R=\log_2(3)/6=0.2642, \delta=1/6)$, which satisfy $R< 1-H_2(\delta)$, hence they lie below the GV curve (see \eqref{GVbound} below). We use this code to illustrate the
three spoiling operations mentioned above.

\begin{itemize}
\item Throughout the entire set of
28 languages considered in \cite{LoGua}, the first two parameters are set to the same value $1$,
hence for the purpose of comparative analysis within this family, we can regard a code like
the above as a twice spoiled code $C=C'\star_1 f_1=(C''\star_2 f_2) \star_1 f_1$ where both
$f_1$ and $f_2$ are constant equal to $1$ and $C''\subset \F_2^4$ is the code obtained from 
the above by canceling the first two letters in each code word.  
\item Conversely, we have $C''=C'\star_2$
and $C'=C\star_1$, in terms of the second spoiling operation described above. 
\item To illustrate
the third spoiling operation, one can see, for instance, 
that $C(0,4)=\{ \ell_1, \ell_3 \}$, while $C(1,6)=\{ \ell_2, \ell_3 \}$.
\end{itemize}

\smallskip

\subsection{The asymptotic bound}

The spoiling operations on codes were used in \cite{Man} to prove the existence of an 
{\em asymptotic bound} in the space of code parameters $(\delta,R)$, see also \cite{Man2},
\cite{ManMar} and \cite{ManMar2} for more detailed properties of the asymptotic bound.

\smallskip

Let $\cV_q \subset [0,1]^2\cap \Q^2$ denote the space of code parameters $(\delta,R)$
of codes $C\subset \F_q^n$ and let $\cU_q$ be the set of all limit points
of $\cV_q$. The set $\cU_q$ is characterized in \cite{Man} as
$$ \cU_q=\{ (\delta,R) \in [0,1]^2\, |\, R\leq \alpha_q(\delta) \} $$
for a continuous, monotonically decreasing function $\alpha_q(\delta)$ (the asymptotic bound).
Moreover, code parameters lying in $\cU_q$ are realized with infinite multiplicity, while
code points in $\cV_q\setminus (\cV_q\cap \cU_q)$ have finite multiplicity and 
correspond to the {\em isolated codes}, see \cite{Man}, \cite{ManMar2}.

\smallskip

Codes lying above the asymptotic bound are codes which have extremely good
transmission rate and relative minimum distance, hence very desirable from the
coding theory perspective. The fact that the corresponding code parameters are not
limit points of other code parameters and only have finite multiplicity reflect the fact 
that such codes are very difficult to reach or approximate. Isolated codes are known
to arise from algebro-geometric constructions, \cite{TsfaVlaZi}, \cite{VlaDri}.

\smallskip

Relatively little is known about the asymptotic bound: the question of the
computability of the function $\alpha_q(\delta)$ was recently addressed in
\cite{Man2} and the relation to Kolmogorov complexity was investigated in
\cite{ManMar2}. There are explicit upper and lower bounds for the 
function $\alpha_q(\delta)$, see \cite{TsfaVla}, including the Plotkin bound
\begin{equation}\label{Plotkin}
 \alpha_q(\delta)=0, \ \  \text{ for } \ \  \delta\geq \frac{q-1}{q} ;
\end{equation}
the singleton bound, which implies that $R=\alpha_q(\delta)$ lies below the
line $R+\delta =1$; the Hamming bound
\begin{equation}\label{Hammingbound}
\alpha_q(\delta) \leq 1 - H_q(\frac{\delta}{2}),
\end{equation}
where $H_q(x)$ is the $q$-ary Shannon entropy
$$ x\, \log_q(q-1) - x \log_q(x) - (1-x) \log_q(1-x) $$
which is the usual Shannon entropy for $q=2$,
\begin{equation}\label{Shannon}
H_2(x) = - x \log_2(x) - (1-x) \log_2(1-x).
\end{equation}
One also has a lower bound given by the Gilbert--Varshamov bound
\begin{equation}\label{GVbound}
\alpha_q(\delta) \geq 
1 - H_q(\delta) 
\end{equation}

\smallskip

The Gilbert--Varshamov curve can be characterized in terms of the
behavior of sufficiently random codes, in the sense of the Shannon
Random Code Ensemble, see \cite{BaFo}, \cite{CoGo}, 
while the asymptotic bound can be characterized in terms of
Kolmogorov complexity, see \cite{ManMar2}.

\medskip

\subsection{Code parameters of language families}

{}From the coding theory viewpoint, it is natural to ask whether
there are codes $C$, formed out of a choice of a collection of
natural languages and their syntactic parameters, whose
code parameters lie above the asymptotic bound curve
$R=\alpha_2(\delta)$.

\smallskip

For instance, a code $C$ whose code parameters violate the
Plotkin bound \eqref{Plotkin} must be an isolated code above
the asymptotic bound. This means constructing a code $C$
with $\delta \geq 1/2$, that is, such that any pair of code words
$w\neq w' \in C$ differ by at least half of the parameters. A direct
examination of the list of parameters in Table A of \cite{LoGua}
and Figure 7 of \cite{LGSBC} shows that it is very difficult to find,
within the same historic linguistic family (e.g. the Indo--European
family) pairs of languages $L_1$, $L_2$ with $\delta_H(L_1,L_2)\geq 1/2$.
For example, among the syntactic relative distances listed in Figure 7 of
\cite{LGSBC} one finds only the pair $({\rm Farsi}, {\rm Romanian})$ with
a relative distance of $0.5$. Other pairs come close to this value, for
example Farsi and French have a relative distance of $0.483$, but
French and Romanian only differ by $0.162$.

\smallskip

One has better chances of obtaining codes above the asymptotic
bound if one compares languages that are not so closely related 
at the historical level. 

\begin{ex}\label{L123}{\rm 
Consider the set
$C=\{ L_1, L_2, L_3 \}$ with languages $L_1=$ Arabic, $L_2=$ Wolof,
and $L_3 =$ Basque. We exclude from the list of
Table A of \cite{LoGua} all those parameters that are entailed and made
irrelevant by some other parameter in at least one of these three chosen languages.
This gives us a list of 25 remaining parameters, which are those numbered
as 1--5, 7, 10, 20--21, 25, 27--29, 31--32, 34, 37, 42, 50--53, 55--57 in \cite{LoGua},
and the following three code words:

{\small
\begin{center}
\begin{tabular}{|c||c|c|c|c|c|c|c|c|c|c|c|c|c|c|c|c|c|c|c|c|c|c|c|c|c|}
\hline
$L_1$ & 1 & 1 & 1 &1 & 1 & 1 & 0 & 1 & 0 & 1 & 0 & 1 & 0 & 1 & 1 & 1 & 1 & 1 & 1 & 0 & 1 & 0 & 0 & 0 & 0 \\ \hline
$L_2$ & 1 & 1 & 1 &0 & 0 & 1 & 1 & 0 & 1 & 0 & 1 & 0 & 0 & 1 & 0 & 1 & 1 & 0 & 0 & 1 & 1 & 1 & 1 & 1 & 1 \\ \hline
$L_3$ & 1 & 1 & 0 &1 & 0 & 0 & 1 & 0 & 0 & 0 & 1 & 1 & 1 & 0 & 1 & 1 & 0 & 1 & 1 & 1 & 1 & 1 & 1 & 0 & 0 \\
\hline
\end{tabular}
\end{center}
}
}
\end{ex}

\smallskip

This example, although very simple and quite artificial in the choice of languages,
already suffices to produce a code $C$ that lies above the asymptotic bound. In fact, we have
$d_H(L_1,L_2)=16$, $d_H(L_2,L_3)=13$ and $d_H(L_1,L_3)=13$, so that
$\delta=0.52$. Since $R>0$, the code point $(\delta,R)$ violates the Plotkin bound,
hence it lies above the asymptotic bound.

\smallskip

It would be more interesting to find a code $C$ consisting of languages belonging
to the same historical-linguistic family (outside of the Indo--European group), that
lies above the asymptotic bound. Such examples would correspond to linguistic
families that exhibit a very strong variability of the syntactic parameters, in a way that
is quantifiable through the properties of $C$ as a code.

\smallskip

If one considers the 22 Indo-European languages in \cite{LoGua} with their
parameters, one obtains a code $C$ that is below the Gilbert--Varshamov line, hence
below the asymptotic bound by
\eqref{GVbound}. A few other examples, taken from other non Indo-European 
historical-linguistic families, computed using those parameters reported in the SSWL database
(for example the set of Malayo--Polynesian languages currently recorded in SSWL)
also give codes whose code parameters lie below the Gilbert--Varshamov curve. 
One can conjecture that any code $C$ constructed out of natural languages
belonging to the same historical-linguistic family will be below the asymptotic
bound (or perhaps below the GV bound), which would provide a quantitative
bound on the possible spread of syntactic parameters within a historical family,
given the size of the family. Examples like the simple one constructed above,
using languages not belonging to the same historical family show that, to the
contrary, across different historical families one encounters a greater variability of
syntactic parameters. To our knowledge, no systematic study of parameter variability from this
coding theory perspective has been implemented so far.

\medskip

\subsection{Entailment and dependency of parameters}\label{entail}

In the discussion above we did not incorporate in our model the fact
that certain syntactic parameters can entail other parameters in such
a way that one particular value of one of the parameters renders
another parameter irrelevant or not defined, see the discussion in
\S 5.3.2 of \cite{LoGua}. 

One possible way to alter the previous construction to account for
these phenomena is to consider the codes $C$ associated to
families of languages as codes in $\F_3^n$, where $n$ is the
number of parameters, as before, and the set of values is now
given by $\{-1,0,+1\}=\F_3$, with $\pm 1$ corresponding to the
binary values of the parameters that are set for a given language
and value $0$ assigned to those parameters that are made
irrelevant for the given language, by entailment from other
parameters, or are not defined.  This allows us to consider the
full range of parameters used in \cite{LoGua} and \cite{LGSBC}.
We revisit Example \ref{L123} considered above.

\begin{ex}\label{L123bis}{\rm Let $C=\{ L_1, L_2, L_3 \}$ be
the code obtained from the languages $L_1=$ Arabic, $L_2=$ Wolof,
and $L_3 =$ Basque, as a code in $\F_3^n$
with $n=63$, using the entire list of parameters in \cite{LoGua}.
The code parameters $(R=0.0252,\delta=0.4643)$ of this code
no longer violate the Plotkin bound. In fact, the parameters satisfy $R< 1-H_3(\delta)$
so the code $C$ now also lies below the GV bound.  }
\end{ex}

Thus, the effect
of including the entailed syntactic parameters in the comparison spoils the
code parameters enough that they fall in the area below the GV bound.

\smallskip

Notice that what we propose here is different from the counting used in \cite{LoGua},
where the relative distances $\delta_H(L_1,L_2)$ are normalized with respect to the
number of non-zero parameters (which therefore varies with the choice of the pair
$(L_1,L_2)$) rather than the total number $n$ of parameters. While this has the desired
effect of getting rid of insignificant parameters that spoil the code, it has the undesirable
property of producing codes with code words of varying lengths, while counting only
those parameters that have no zero-values over the entire family of languages, as
in Example \ref{L123} avoids this problem. Adapting the coding theory results about
the asymptotic bound to codes with words of variable length may be desirable for
other reasons as well, but it will require an investigation beyond the scope of 
the present paper.

\smallskip

More generally, there are various kinds of dependencies among
syntactic parameters. Some sets of hierarchical relations 
are discussed, for instance, in \cite{Baker}.

By the spoiling operations $C\star_i f$ of codes described above, we know
that if some of the syntactic parameters considered are functions of other
parameters, the resulting code parameters of $C\star_i f$ are worse than
the parameters of the code $C$ where only independent parameters were
considered. 

Part of the reason why code parameters of groups of
languages in the family analyzed in \cite{LoGua} end up in the region
below the asymptotic and the GV bound may be an artifact of the presence
of dependences among the chosen 63 syntactic parameters. From the
coding theory perspective, the parametric comparison method works best
on a smaller set of independent parameters than on a larger set that
includes several dependencies.

\medskip

\section{Conclusions}

We proposed an approach to the linguistic parametric comparison method
of \cite{Longo}, \cite{Longo2} via the mathematical theory of error-correcting
codes, by assigning a code to a family of languages to be analyzed with
the PCM, and investigating its position in the space of code parameters,
with respect to the asymptotic bound and the GV bound. We have shown that
there are examples of languages not belonging to the same
historical-linguistic family that yield isolated codes above the asymptotic bound,
while languages belonging to the same historical-linguistic family appears to
give rise to codes below the bound, though a more systematic analysis would
be needed to map code parameters of different language groups.

We have also shown that, from these coding theory perspective, it is preferable
to exclude from the PCM all those parameters that are entailed and made
irrelevant by other parameters, as those spoil the properties of the resulting 
code and produce code parameters that are artificially low with respect
to the asymptotic bound and the GV bound.

\subsection{Questions}

The approach to the PCM based on error-correcting codes proposed here
suggests a few new linguistic questions that may be suitable for treatment
with coding theory methods:
\begin{enumerate}
\item Do languages belonging to the same historical-linguistic family
always yield codes below the asymptotic bound or the GV bound? How often does the
same happen across different linguistic families? How much can code
parameters be improved by eliminating spoiling effects caused by
dependencies and entailment of syntactic parameters?
\item Codes near the GV curve are typically coming from the Shannon
Random Code Ensemble, where code words and letters of code words
behave like independent random variables, see \cite{BaFo}, \cite{CoGo}.
Are there families of languages whose associated codes are located
near the GV bound? Do their syntactic parameters mimic the random
behavior?
\item The asymptotic bound for error-correcting codes was related in
\cite{ManMar2} to Kolmogorov complexity. Is there a suitable complexity
measure associated to a family of natural languages that would relate to the
position of the resulting code above or below the asymptotic bound?
\item Codes and the asymptotic bound in the space of code parameters
were recently studied using methods from quantum statistical mechanics,
operator algebra and fractal geometry, \cite{ManMar}, \cite{MaPe}.
Can some of these mathematical methods be employed in the
linguistic parametric comparison method?
\end{enumerate}

\end{document}